\documentclass[letterpaper, 10 pt, conference]{ieeeconf}  

\IEEEoverridecommandlockouts                              

\usepackage{graphicx} 
\usepackage{amsmath} 
\usepackage{amssymb}  
\usepackage{multirow}
\usepackage{multicol}
\usepackage[T1]{fontenc}
\usepackage{paralist}
\usepackage[hidelinks]{hyperref}
\let\labelindent\relax
\usepackage[inline]{enumitem}
\newcommand{\eg}{\emph{e.g.}, }
\newcommand{\ie}{\emph{i.e.}, } 
\newcommand{\etal}{\emph{et al.} } 
\usepackage{diagbox}
\usepackage{textcomp}
\usepackage{breqn}
\usepackage{booktabs}
\usepackage[dvipsnames]{xcolor}

\definecolor{mypurple}{RGB}{115,51,128}
\definecolor{myred}{RGB}{250,50,83}
\definecolor{mygreen}{RGB}{36,179,83}

\title{\LARGE \bf 
Semantically-Aware Diver Activity Recognition Framework for\\ Effective Underwater Multi-Human-Robot Collaboration
}

\author{Sadman Sakib Enan$^{1}$ and Junaed Sattar$^{2}$
\thanks{$^{*}$This work was supported in part by the National Science Foundation Grant IIS-\#2220956.
The authors were with the department of Computer Science \& Engineering and the Minnesota Robotics Institute, University of Minnesota, MN, USA at the time this work was conducted. Email: {\tt\small $^{1}$sadmansakib.enan@gmail.com, $^{2}$junaed@umn.edu}}
}

\begin{document}

\maketitle
\thispagestyle{empty}
\pagestyle{empty}

\begin{abstract}

Effective multi-human-robot collaboration is essential for expanding human-led operations in the challenging and high-risk underwater environment. 
For autonomous underwater vehicles (AUVs) to become true teammates, they must be able to comprehend their surroundings and recognize a diver's activities to offer assistance and ensure safety. 
Towards this goal, we introduce DAR-Net, a novel transformer-based framework that analyzes complex underwater scenes to classify diver activities. 
Our contribution lies in a semantically guided learning formulation that couples transformer-based temporal reasoning with pixel-level scene supervision. This multi-loss training strategy explicitly aligns global activity recognition with local human–robot interaction semantics, which is particularly critical in low-visibility underwater conditions.
To address the significant challenge of data scarcity in this domain, we present the first-ever Underwater Diver Activity (UDA) dataset, a foundational resource containing over $2,600$ annotated images with pixel-level masks. 
Through rigorous experimental evaluations in a controlled environment, we demonstrate that DAR-Net achieves promising accuracy in recognizing six distinct diver activities, outperforming state-of-the-art models. 
While this dataset provides a crucial baseline, our work serves as a pioneering step, laying the groundwork for future research and facilitating the development of more intelligent, collaborative underwater robotic systems.	
	
\end{abstract}

\section{Introduction}
The use of Autonomous Underwater Vehicles (AUVs) has seen significant expansion across a broad range of tasks, such as environmental monitoring~\cite{girdhar2023curee}, mapping~\cite{wang2023realtime}, submarine cable and wreckage inspection~\cite{lickliter2023monitoring}, and search-and-rescue operations~\cite{wu2022reinforcement}. 
This growth is primarily attributed to advancements in on-board computational power and enhanced Underwater Human-Robot Interaction (UHRI) capabilities, which enable AUVs to interact effectively with human divers without assistance from a topside operator (\eg\cite{islam2019understanding}).
Recent research~\cite{enan2022visual} allows AUVs to determine diver attentiveness and plan trajectories for interaction if needed, making them more powerful for collaborative missions with human divers. 
For this collaboration to be truly effective, an AUV must be more than just a tool; it needs to be an intelligent partner capable of understanding its surroundings and its human teammates (\ie \emph{dive buddies}). 
A critical component of this is the ability to recognize a diver's current activity and the significance of that activity. 
For instance, during a sensitive task like a rescue mission, the AUV must not disrupt the diver's focus, allowing them to concentrate fully on their responsibilities without unnecessary interference. 
Therefore, equipping an AUV with the capability to recognize a diver's activity is not merely an enhancement--it is a necessity for informed decision-making and safe, efficient collaboration.

\begin{figure}[t]
    \vspace{2mm}
    \centering    
    \includegraphics[width=0.9\linewidth]{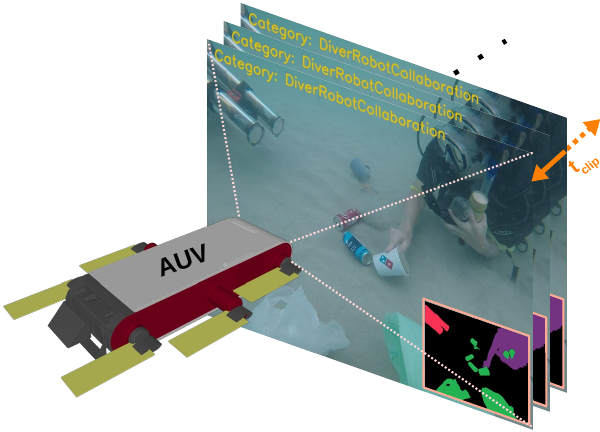}
    \caption{Demonstration of the proposed diver activity recognition framework, making inference on real-world underwater scenarios involving multiple divers and robots. The proposed method is able to learn highly discriminative spatio-temporal features from underwater video clips by focusing on important elements within the scene (as shown in the inset), such as divers, robots, and objects of interest, and their interactions with each other.}
    \label{fig:intro}
    \vspace{-4mm}
\end{figure}

While significant progress has been made in terrestrial activity recognition (\eg\cite{xia2020lstm-cnn,gavrilyuk2020actor-transformers}), the underwater environment presents unique and formidable challenges. 
Data collection is difficult due to high pressure, low visibility, extreme temperatures, and unpredictable currents~\cite{Wei2022_underwaterDataChallenges}, leading to a near-complete absence of large-scale underwater diver activity datasets, hampering the design of deep-learned recognition methods. 

To address this critical gap, we introduce the first-ever \emph{Underwater Diver Activity (UDA)} dataset, which contains over $2600$ semantically segmented images with annotations for divers, robots, and objects of interest in a multi-human-robot collaborative setting.
We also propose a novel, end-to-end framework named \emph{DAR-Net} (\emph{Diver Activity Recognition Network}) to analyze and understand diver activities. 
This transformer-based architecture is designed to extract highly discriminative spatio-temporal features from underwater scenes. 
Crucially, our approach uses supervision from scene semantics to train the model to focus on the most important elements -- the divers, robots, and objects -- instead of irrelevant background noise (\eg\cite{islam2022svam}). 
This enables the AUV to take proactive, informed actions, such as joining a team to remove trash after recognizing their collaborative activity (see Fig.~\ref{fig:intro}). 
Through comprehensive experimental evaluations, we demonstrate that our framework achieves a promising accuracy and outperforms existing state-of-the-art activity recognition models.

We make the following contributions in this paper:
\begin{enumerate}
    \item We propose an end-to-end transformer-based deep network called DAR-Net to analyze and classify different diver activities in underwater multi-human-robot collaborative scenes. The training pipeline includes a multi-loss objective function that prioritizes important regions to learn from, instead of learning from the whole image.
    \item Additionally, we present UDA, the first-ever Underwater Diver Activity dataset that includes $2640$ annotated underwater multi-human-robot collaborative scenes divided into $6$ diver activity categories. The data were collected from several closed-water robot trials and contain pixel-level annotations for divers, robots, and objects of interest.
    \item Furthermore, we conduct both quantitative and qualitative experiments, demonstrating that the proposed framework derives significant benefits from incorporating additional supervision from semantic labels. This enables the model to learn highly discriminative spatio-temporal features essential for recognizing different diver activities.
\end{enumerate}
\section{Related Work}
While significant advancements have been made in autonomous robotics for various underwater tasks, the field of UHRI remains an emerging and challenging domain. 
A critical bottleneck to realizing truly effective collaboration is the lack of a robust system for diver activity recognition. 
While research in terrestrial activity recognition is extensive, the unique challenges of the underwater environment -- such as low visibility, unpredictable lighting, and the complex interactions of divers with their surroundings -- have left this area largely unexplored.


Human Activity Recognition (HAR) is an active research area within computer vision and robotics, with research efforts extending over several decades~\cite{lara2013asurvey,vrigkas2015review,ke2013review}. 
Sensor-based HAR has been a cornerstone in activity recognition research, owing to its ubiquity and ease of data collection~\cite{serpush2022wearable}. 
Early approaches (\eg\cite{wang2011recognizing}) utilized wearable sensors and traditional machine learning models like Hidden Markov Models (HMMs) and Support Vector Machines (SVMs). 
Advancements in deep learning have led to more sophisticated models (\eg\cite{jiang2015human}), such as Convolutional Neural Networks (CNNs), that learn features directly from raw sensor data.

Vision-based HAR (\eg\cite{girish2020understanding}) has also gained significant traction due to the proliferation of cameras. 
Early work focused on handcrafted features (\eg\cite{huang2011human,su2013human}), while more recent deep learning techniques (\eg\cite{dobhal2015human}) have revolutionized the field. 
Researchers have explored architectures like two-stream CNNs that process both spatial and temporal information from video frames, achieving state-of-the-art (SOTA) performance in action recognition tasks~\cite{simonyan2014two,feichtenhofer2019slowfast}. 
Early approaches typically involved handcrafted feature extraction from video sequences~\cite{robertson2006ageneral,zelnikmanor2001event}, followed by classification using methods, such as SVMs or decision trees. 
With the advent of deep learning, however, researchers have developed end-to-end architectures for video-based HAR (\eg\cite{tran2015learning,carreira2017quo})
Hara~\etal\cite{hara2017learning} have also proposed to use 3D kernels in CNNs to extract spatio-temporal features from videos for activity recognition. 
Although 3D CNNs are typically susceptible to overfitting due to their large number of parameters, the use of large-scale activity datasets (\eg UCF-101~\cite{soomro2012ucf101}, Sports-1M~\cite{karpathy2014largescale}, ActivityNet~\cite{heilbron2015activitynet}, Kinetics~\cite{kay2017kinetics}) for training has mitigated the issue. 
In contrast, Wu~\etal\cite{wu2019longterm} have showed that augmenting 3D CNNs with a long-term feature bank can yield SOTA results in activity recognition task. 
Furthermore, advancements in Large Language Models (LLMs) have helped create robust HAR methods~\cite{girdhar2019video,kalfaoglu2020late}.   

On the contrary, HAR techniques for underwater domain have received considerably less attention. 
While there have been a few research endeavors addressing issues, such as diver motion prediction in the context of UHRI~\cite{hu2020underwater,yang2023dare,delhaye2022automatic}, non-human motion prediction~\cite{maaloy2019spatio,enan2022robotic}, and monitoring divers to ensure safety~\cite{goodfellow2015divernet,nadj2019towards}, none of these specifically tackle the problem of diver activity recognition. 
This is primarily due to the lack of large-scale human activity recognition datasets tailored specifically for underwater environments. 
There are a few diver dataset available in the literature (\eg \cite{gomezchavez2019caddy}), however, they do not include data of divers actively engaged in different activities or tasks, specifically collaborating with AUVs. 
To this end, we propose to formulate the first-ever UDA dataset involving multiple divers and AUVs, so that it can be used to supervise the learning and validation of diver activity recognition frameworks.
\section{UDA Dataset}
\begin{figure}[t]
    \vspace{2mm}
    \centering    
    \includegraphics[width=1\linewidth]{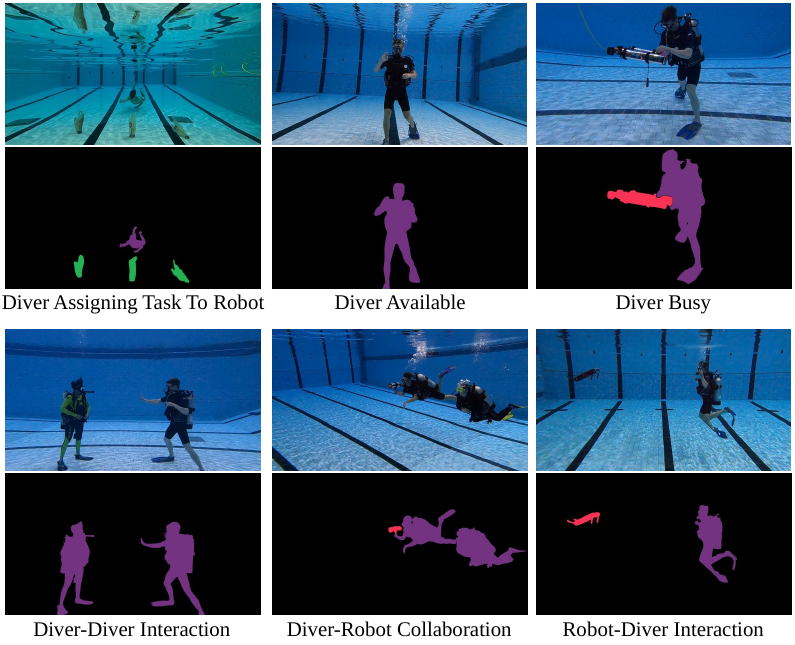} 
    \caption{A few sample images and their semantic labels from the proposed UDA dataset. In our dataset, divers, robots, and objects of interest are annotated with \textcolor{mypurple}{purple}, \textcolor{myred}{red}, and \textcolor{mygreen}{green} colors, respectively.}
    \label{fig:dataset}
    \vspace{-4mm}
\end{figure}
The lack of publicly available, large-scale datasets for underwater human-robot activity has long been a major impediment to research in this field. 
To address this critical gap, we have curated the UDA dataset. 
This dataset is the first of its kind, meticulously compiled from real-world human-robot collaborative trials conducted in closed-water environments and specifically designed to support the development of diver activity recognition frameworks. 
We collect the data as video clips having resolutions of $1920\times 1080$ pixels, using GoPros~\cite{gopro8}. 
All data are captured in a closed-water pool environment with controlled lighting and visibility. The scenes include between one and three divers, one to two robots, and one to three task-specific objects. Each clip is approximately three seconds long and captures diver poses arising naturally from real task execution.
These categories were chosen based on our extensive research and engagement with aquatic professionals to represent key interactions between multiple divers and robots. 
We chose these six categories as representative activities to demonstrate the efficacy of our proposed framework for the classification task. 
The activities are defined as follows:
\begin{enumerate}
    \item \textit{Diver Assigning Task to Robot:} A diver gives instructions to a robot, often without the robot being in the scene.
    \item \textit{Diver Available:} A diver is present but not actively engaged, indicating an opportunity for interaction. 
    \item \textit{Diver Busy:} A diver is occupied with a task and is not available for interaction, even if a robot is in the scene.
    \item \textit{Diver-Diver Interaction:} Two divers are communicating with each other.
    \item \textit{Diver-Robot Collaboration:} Divers and robots are actively working together on a shared task.
    \item \textit{Robot-Diver Interaction:} A diver and robot are engaged in non-verbal communication, separate from active task execution.
\end{enumerate}

Each image in the dataset has been carefully annotated at the pixel level to capture the scene's semantics. 
Using the Segmentation Anything Model (SAM)~\cite{kirillov2023segment} as a base, we created detailed segmentation masks for \emph{divers}, \emph{robots}, and \emph{objects of interest}, which were manually verified and corrected to ensure accurate boundaries.
These pixel-level annotations are crucial, as they allow our framework to learn from and focus on the most relevant elements within a scene.

Unlike existing underwater datasets that primarily capture isolated diver presence or motion, UDA focuses on task-driven, multi-human–robot interactions. Each activity category includes visually diverse instances with varying numbers of divers, robot proximity, interaction patterns, and object configurations, providing meaningful intra-class variability for activity recognition.

The UDA dataset is a cornerstone of our research, enabling the training of robust deep-learning models for a task previously considered infeasible. 
We make this dataset publicly available at~\url{https://irvlab.cs.umn.edu/uda}, to accelerate future research in the critical domain of underwater human-robot collaboration.
%
\section{Diver Action Recognition Framework}
Our proposed DAR-Net (Diver Activity Recognition Network) is an end-to-end framework designed to process underwater video clips and extract robust spatio-temporal features to recognize various diver activities. 
The core of our approach lies in a unique multi-loss objective function that minimizes both a classification loss and a segmentation loss, enabling the model to learn from both global and local context simultaneously.

\begin{figure*}[t]
    \vspace{2mm}
    \centering    
    \includegraphics[width=0.9\linewidth]{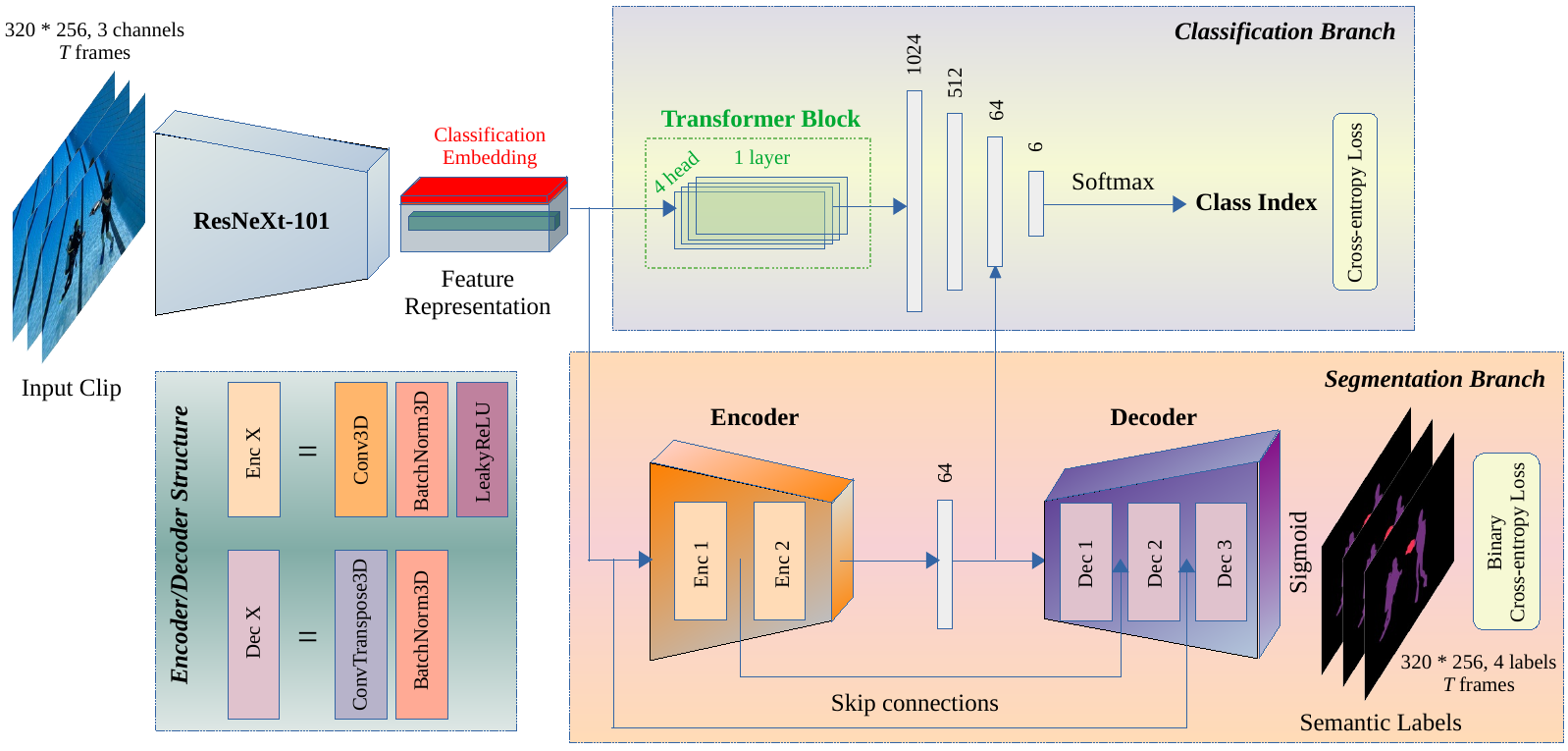} 
    \caption{An overview of the network architecture of DAR-Net. It takes an underwater diver activity video clip as input and extracts highly discriminative spatio-temporal features by incorporating additional supervision from scene semantics. Intermediary skip connections are used to avoid overfitting.}
    \label{fig:model}
    \vspace{-4mm}
\end{figure*}
\subsection{Feature Extraction}
To effectively extract rich, discriminative features from the underwater environment, we use a ResNeXt-101~\cite{xie2017aggregated} network as the backbone of our model. 
This architecture is renowned for its high modularity and ability to capture complex features, and has shown good performance in underwater applications (\eg\cite{enan2022robotic}).
ResNeXt employs a ``\emph{split-transform-aggregate}'' strategy, which creates a homogeneous, multi-branch structure that is highly efficient for learning. 
By increasing the network's cardinality (the size of its transformations), we achieve a significant improvement in performance over traditional approaches that simply increase depth or width.

Following recent advancements in video action recognition~\cite{girdhar2019video,kalfaoglu2020late}, we further enhance our feature representation by incorporating both positional and classification encodings. 
These encodings, initialized with a normal distribution $\mathcal{N}(0, 0.02^2)$, ensure that each feature location retains its positional information, which is crucial for accurately classifying activities. 
This enriched feature representation is then passed to two distinct branches for further processing (see Fig.~\ref{fig:model}): the \emph{Classification Branch} and the \emph{Segmentation Branch}, which we describe below.

\begin{figure}[t]
    \vspace{2mm}
    \centering    
    \includegraphics[width=1\linewidth]{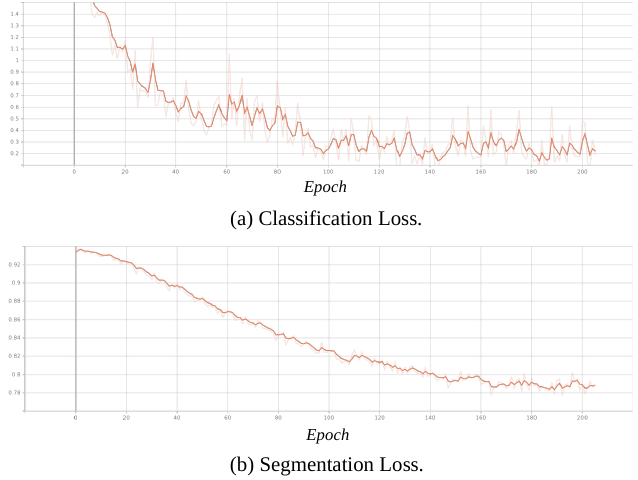} 
    \caption{The training performance of DAR-Net. Note the convergence in both the classification and segmentation validation loss graphs.}
    \label{fig:training}
    \vspace{-4mm}
\end{figure}

\subsection{Objective Function Formulation}
The core of DAR-Net's learning process is our multi-loss objective function, which allows the model to analyze and determine diver activities in a robust manner. 
We design two separate processing branches for the spatio-temporal features:
\begin{compactenum}
\item \emph{The Classification Branch}: Based on a transformer~\cite{girdhar2019video} architecture, this branch is designed to focus on the most important temporal regions of the features, effectively capturing the global context of the diver's actions. 
The final output of this branch is supervised by the class labels and trained with a standard cross-entropy loss function.
\item \emph{The Segmentation Branch}: This branch, built on an encoder-decoder architecture, is used to leverage spatial scene semantics, thereby capturing the local context. This branch is supervised by the semantic labels and trained with a binary cross-entropy loss function.
\end{compactenum}

This hybrid learning strategy, guided by both global classification and local semantic information, facilitates the learning of robust features for the classification task. 
Given $x_{n,y_n}$ as the unnormalized logit for the diver activity category $y_n$, where $n$ refers to the $n$-th sample from the minibatch, we define the classification cross-entropy loss $\mathcal{L}_{class}$ and the segmentation binary cross-entropy loss $\mathcal{L}_{seg}$ functions as follows. 
 
\begin{equation*}
    \mathcal{L}_{class} = -\sum_{n=1}^{N}{\text{log}\frac{\text{exp}(x_{n,y_n})}{\sum_{i=1}^{\tau}\text{exp}(x_{n,i})}}y_n
\end{equation*}
\begin{equation*}
    \mathcal{L}_{seg} = -\sum_{n=1}^{N}{[y_n \text{log}x_{n,y_n} + (1-y_n) \text{log}(1-x_{n,y_n})]}
\end{equation*}
where $N$ is the batch size, and $\tau=6$ is the number of activity categories.

The overall training is performed by minimizing a combined multi-loss objective function:
\begin{equation}
	\mathcal{L} = \alpha \mathcal{L}_{class} + \beta \mathcal{L}_{seg} \label{eq:total_loss}
\end{equation}
The weights $\alpha$ and $\beta$ are set as trainable parameters, allowing the model to dynamically adjust the importance of each loss during training.

\subsection{Implementation Details}
We implemented our framework using the PyTorch library~\cite{paszke2019pytorch}.
Although the transformer backbone follows from our prior work~\cite{enan2022robotic}, the primary contribution of this work is the integrated injection and joint optimization of semantic supervision with activity classification.
For the segmentation branch, we employed an encoder-decoder architecture to learn from the scene semantics and fed the encoded information into the classification branch. 
Intermediary skip connections were used within the segmentation branch to mitigate the vanishing gradient problem, further enhancing the model's performance and stability.
The individual elements of each encoder and decoder block is shown in Fig.~\ref{fig:model}.
The $\alpha$ and $\beta$ values in the multi-loss objective function (Eq.~\ref{eq:total_loss}) were set as trainable parameters. 
We employed various data augmentation techniques~\cite{wang2015towards}, including random cropping, image distortion, and flipping, to increase the robustness of the model.
The model was trained on the UDA dataset using an $80/20$ split for training and validation, respectively, for $200$ epochs on an Nvidia RTX6000 Ada Generation GPU; validation losses were observed to converge during training (Fig.~\ref{fig:training}).
We used a batch size of $4$, a learning rate of $10^{-5}$, and the ADAMW optimizer~\cite{loshchilov2017decoupled} with a momentum of $0.9$. 
The input data were resized to a spatial resolution of $320\times256$ pixels and processed in $64$-frame video chunks, which represent approximately $3$ seconds of video. 
\section{Experimental Evaluations}
This section outlines the process for evaluating the performance of the proposed DAR-Net framework against several SOTA models for diver activity recognition.

\subsection{Evaluation Process and Metrics}
To ensure a fair and rigorous evaluation, we created a test set of $30$ video clips that were distinct from the training and validation data. 
This test set includes five video clips for each of the six diver activity categories. 
To maintain fairness, all SOTA baselines were fully retrained on the UDA dataset using their recommended hyperparameters.

\begin{table}[t]
    \vspace{2mm}
    \centering
    \caption{Diver activity recognition performance* on the test set. Precision, Recall, and F1-Score are computed as weighted averages. The values are in percentages.}
    \begin{tabular}{lcccc}
    \toprule
      \textbf{Method}   &  \textbf{Accuracy} & \textbf{Precision} & \textbf{Recall}  & \textbf{F1-Score}\\
      \midrule
      3DResNet~\cite{hara2017learning} &  $53.33$  & $59.84$  &  $53.33$ & $53.31$ \\
      R(2+1)D~\cite{tran2018acloser} &  $60.00$  & $70.63$  &  $60.00$ & $56.83$ \\
      SlowFast~\cite{feichtenhofer2019slowfast} & $56.67$   & $65.00$  &  $56.67$ & $57.67$\\
      LateTemporal~\cite{kalfaoglu2020late} &  $66.67$  &  $71.25$ & $66.67$  & $65.60$\\
      RRCommNet~\cite{enan2022robotic} &  $60.00$  & $67.64$  & $60.00$  & $61.05$\\
      \textbf{Ours} & \boldmath $73.33$ \unboldmath   & \boldmath $76.90$ \unboldmath & \boldmath $73.33$ \unboldmath & \boldmath $72.17$ \unboldmath \\
      \bottomrule \\
      \multicolumn{5}{l}{*The performance is evaluated based on the classification accuracy.}
    \end{tabular}
    \label{tab:acc_f1}
    \vspace{-6mm}
\end{table}
\begin{figure*}[t]
    \vspace{2mm}
    \centering\includegraphics[width=0.82\linewidth]{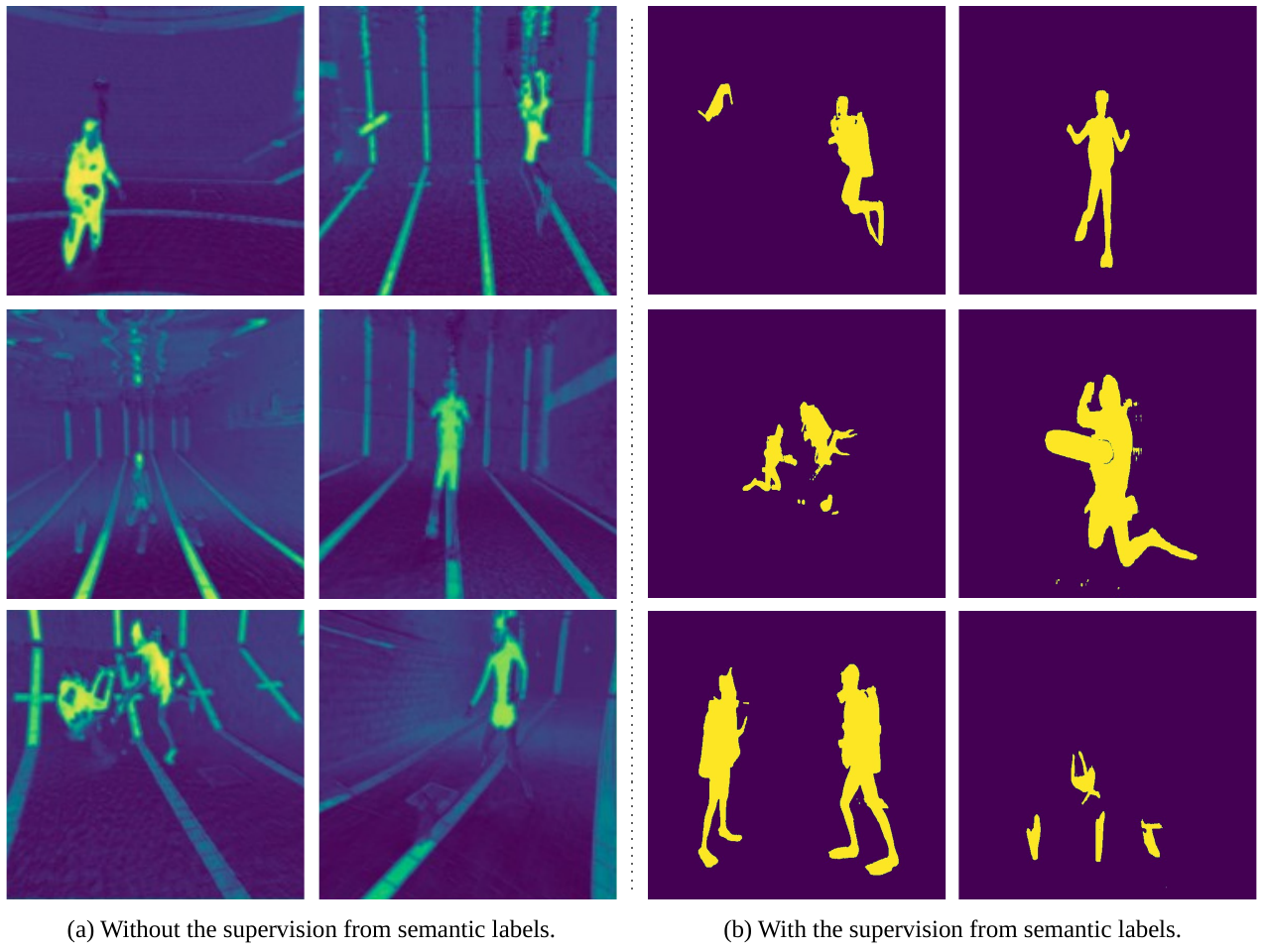} 
    \caption{The effect of using semantic labels during the training of DAR-Net. The inclusion of scene semantics directs the model's focus towards relevant regions in the image for feature learning. In contrast, traditional activity recognition models frequently prioritize irrelevant areas, such as pool lane markings during training. Lighter colors indicate higher attention values. The attention maps on the right are generated by applying binary thresholding to the output of the sigmoid function from the segmentation branch.}
    \label{fig:attention_maps}
    \vspace{-4mm}
\end{figure*}

Each video clip in the test set, representing approximately $3$ seconds of footage, was processed in chunks of $64$ frames. 
These clips, in a four-dimensional RGB tensor format $(64,3,h_{im},w_{im})$ where $h_{im}$ and $w_{im}$ are height and width of the image, respectively, were fed into the trained model to produce classification scores, $\textbf{x}_{pred} = [x^1_{pred}, \dots, x^6_{pred}]^T$. 
A softmax function was then applied to convert these scores into probability scores:

\begin{equation*}
    P(x^i_{pred}) = \frac{\text{exp}(x^i_{pred})}{\sum_{j=1}^{\tau}{\text{exp}(x^j_{pred})}}
\end{equation*}

where $i$ refers to the $i$-th category index and can have values in the range $[1,\tau]$. 
Finally, the predicted category is found by selecting the index with the maximum probability score.

The performance of each model was measured using four key metrics:
\begin{enumerate}
    \item \textit{Accuracy:} The overall correctness of the model's predictions, calculated as $\frac{TP+TN}{TP+TN+FP+FN}$. Here, $TP$ = True Positives, $TN$ = True Negatives, $FP$ = False Positives, and $FN$ = False Negatives. 
    \item \textit{Precision:} Measures the model's ability to avoid false positive predictions, calculated as $\frac{TP}{TP+FP}$.
    This is particularly useful when the cost of false positives is high.
    \item \textit{Recall:} Measures the model's ability to correctly identify all positive samples, calculated as $\frac{TP}{TP+FN}$. 
    Recall is particularly useful when the cost of false negatives is high.
    \item \textit{F1-Score:} Combines precision and recall into a single metric, providing an overall measure of effectiveness for the classification task. 
    It is calculated as $\frac{2\times Precision \times Recall}{Precision + Recall}$. 
    This metric can help assess the model's overall effectiveness in the classification task.
\end{enumerate}

\subsection{Results}
Our experimental results demonstrate that DAR-Net consistently outperforms the SOTA models. 
As shown in Table~\ref{tab:acc_f1}, our framework achieves a classification accuracy of $73.33\%$, which is notably higher than the other models tested. 
This is significant, as activity recognition from video is a complex task, and the SOTA models struggled to accurately classify diver activities. 
The Late Temporal model was the only one that achieves a comparable classification accuracy of $66.67\%$, likely due to its use of transformer blocks and the self-attention mechanism~\cite{vaswani2017attention}.
We note that some SOTA baselines, originally designed for larger datasets, may not fully converge under limited data availability, further underscoring the benefit of semantically guided supervision in data-scarce underwater applications.

The superior performance of DAR-Net is further validated by its high average precision ($76.90\%$), recall, and F1-score. 
This indicates that our framework is highly effective at accurately classifying relevant activities while maintaining low rates of both false positives and false negatives. 

\begin{figure}[t]
    \vspace{2mm}
    \centering    
    \includegraphics[width=1\linewidth]{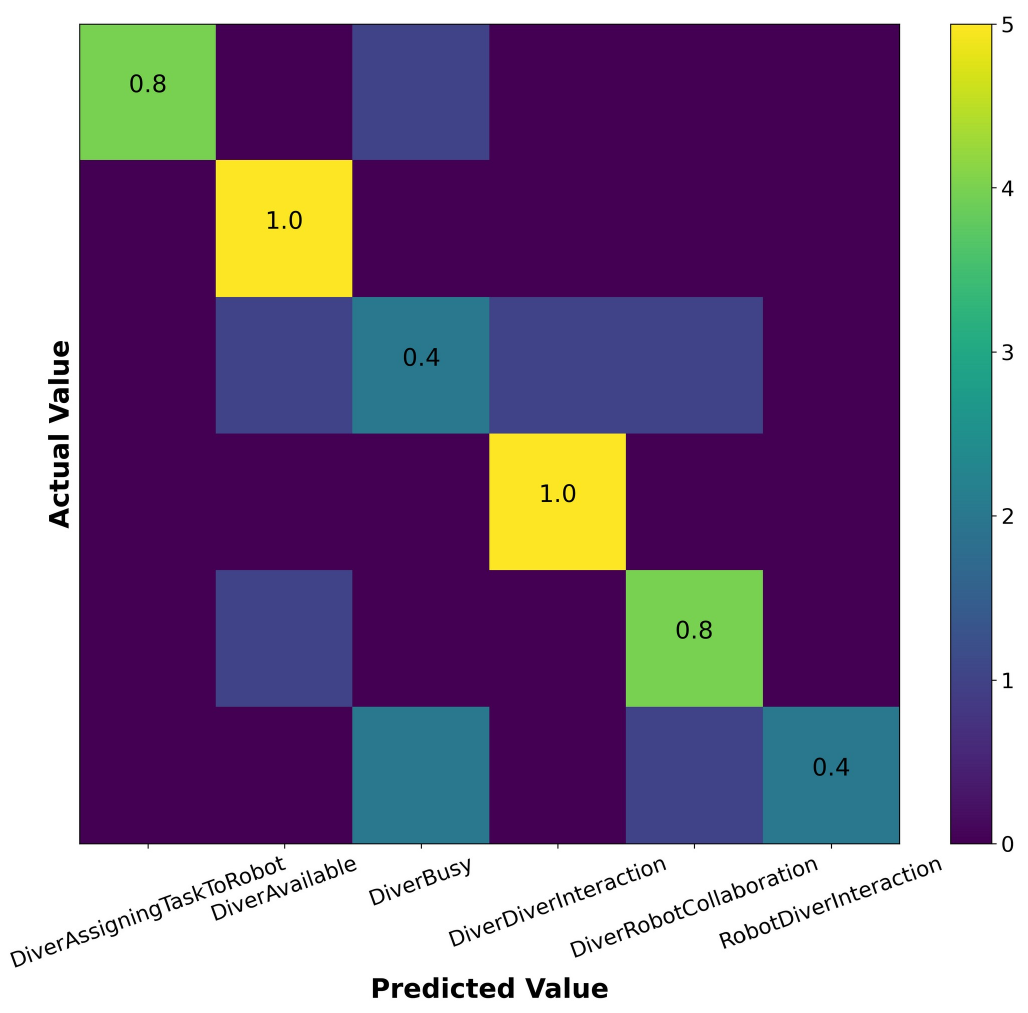} 
    \caption{Confusion matrix for diver activity recognition, computed on the $30$ video clips from our test set. It highlights the robustness of the proposed framework in accurately identifying the majority of diver activity categories. The matrix entries are normalized.}
    \label{fig:cm}
\end{figure}
\begin{figure}[t]
    \centering    
    \includegraphics[width=1\linewidth]{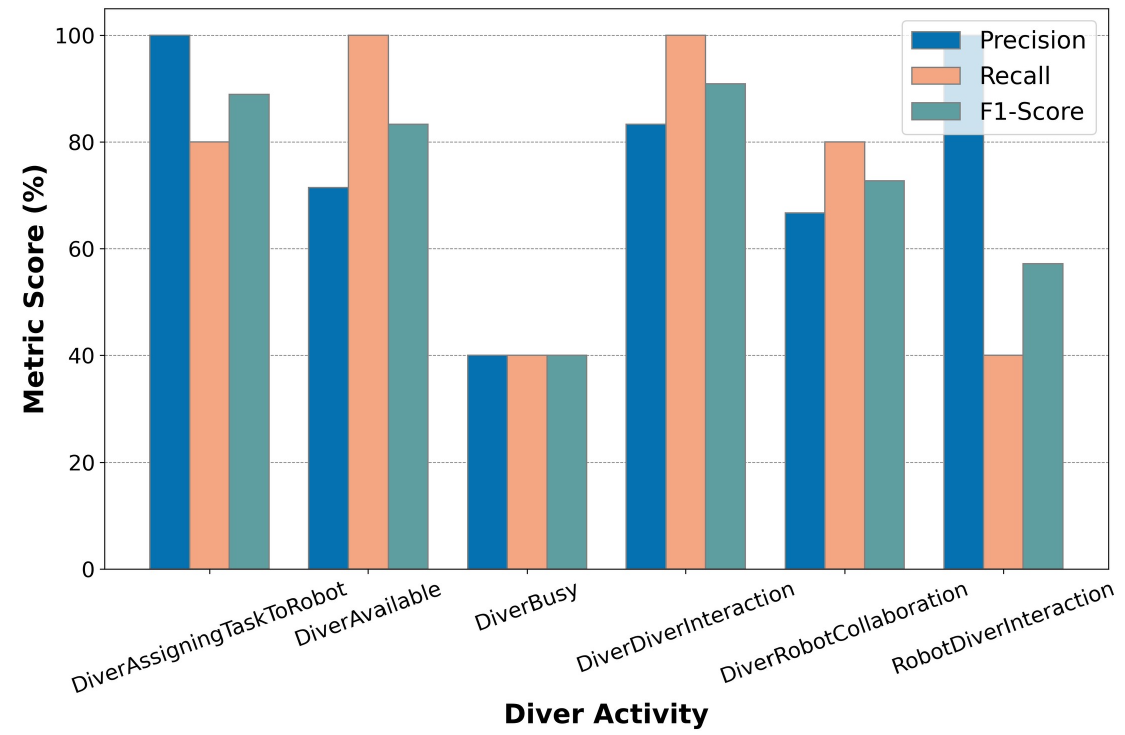} 
    \caption{Per-category Precision, Recall, and F1-Score for the proposed diver activity recognition framework. It indicates a similar trend where the model's performance is notably lower in predicting the \textit{Diver Busy} and \textit{Robot-Diver Interaction} categories.}
    \label{fig:prf1}
    \vspace{-4mm}
\end{figure}   

Furthermore, we conduct an ablation study that isolates the effect of semantic supervision by comparing training with and without segmentation-based supervision under otherwise identical settings.
In Fig.~\ref{fig:attention_maps}, we visualize attention maps from the activity recognition model, highlighting the image regions prioritized during training. 
Fig.~\ref{fig:attention_maps}a illustrates the attention maps obtained from the model trained without semantic supervision.
From the figure, it is evident that the model focuses unnecessarily on image regions irrelevant to determining the underlying diver activity category, such as lane markings on the pool bottom and sides. 
This could contribute to the lower accuracy recorded when scene semantics were not considered during training. 
In contrast, upon integrating scene semantics into the training process of our proposed framework, the model directs attention solely to crucial image regions, such as divers, robots, and objects of interest, as shown in Fig.~\ref{fig:attention_maps}b. 
With DAR-Net's training supervised by semantic labels, the intermediary attention maps naturally prioritize the segmented regions, leading to superior accuracy in identifying diver activity categories. 
The attention maps in Fig.~\ref{fig:attention_maps}b (segmentation masks, in this case) are generated by applying binary thresholding to the output of the sigmoid function from the segmentation branch.

We also show a confusion matrix for the diver activity recognition task (see Fig.~\ref{fig:cm}) by recording the predictions made by DAR-Net on our test set consisting of $30$ video clips of underwater multi-human-robot collaborative scenarios. 
As illustrated in the figure, DAR-Net accurately classifies the majority of different diver activities. 
However, it encounters challenges in identifying certain activities, most notably \textit{Diver Busy} and \textit{Robot-Diver Interaction}. Upon closer examination of video clips from these categories, we observe that both often involve a single diver co-located with a robot and limited explicit gesturing, resulting in shared visual primitives. In several failure cases, the distinction depends on subtle temporal cues (\eg tool manipulation vs. communicative posture), which are weakly represented in short clips and further degraded by turbidity and lighting artifacts.

To further analyze this behavior, we compute the per-category precision, recall, and F1-score and visualize the results in Fig.~\ref{fig:prf1}. A consistent trend emerges, showing lower performance for the aforementioned two categories compared to others. Specifically, for the \textit{Diver Busy} category, the model struggles to detect positive samples, leading to low recall, while also producing false positives, resulting in reduced precision. In contrast, the \textit{Robot-Diver Interaction} category is detected less consistently; however, when identified, the predictions are reliable, as reflected by high precision and low recall. For the remaining activity categories, DAR-Net demonstrates comparatively strong and balanced performance. This observed discrepancy highlights the need for further investigation into temporally disentangling these visually similar activities to reduce misclassification.

\subsection{Limitations and Mitigation}
\label{sec:limitations}
While the proposed work serves as a foundational step toward enabling effective multi-human-robot collaboration in the challenging underwater environment, it is important to acknowledge the inherent limitations of this research. 
A key limitation of this study is the size and scope of the Underwater Diver Activity (UDA) dataset. 
While we have meticulously curated the first-of-its-kind dataset with over $2,600$ annotated images, this size is considered small for a transformer-based network, which typically benefits from larger and more diverse data volumes to demonstrate strong generalization capacity. 
Underwater data collection is a uniquely difficult process due to safety concerns, logistical complexities, and the challenge of obtaining appropriate Institutional Review Board (IRB) approvals for human trials. 
This makes creating a large-scale dataset, comparable to those in terrestrial robotics, an arduous task.
Furthermore, our experiments were conducted in a closed-water environment. 
This focused scope, while necessary for a controlled study, may not fully represent the variability and unpredictability of real-world, open-water conditions. To address these limitations, future work will focus on three key areas:
\begin{compactenum}
	\item \emph{Dataset Expansion}: We will explore advanced data augmentation techniques and synthetic data generation to virtually expand the UDA dataset. This will improve the model's robustness and help it generalize to new scenarios beyond the scope of our initial data collection. We will also collaborate closely with our institutional ethics board to create protocols for approved open-water data collection trials to increase dataset diversity.
	
	\item \emph{Model Robustness}: We will conduct more extensive experiments to show the efficacy of our proposed method. 
	A deeper analysis will be performed on existing categories (and any novel ones we introduce) to identify and decouple the visual cues that lead to confusions among them.
	
	\item \emph{Community Collaboration}: By making the UDA dataset publicly available, we will invite the research community to contribute, expand, and use this resource to accelerate the development of robust and generalizable underwater perception systems.	
\end{compactenum}

\section{Conclusions}

This work presents a significant stride in the field of underwater human-robot collaboration by introducing a novel, end-to-end framework for diver activity recognition. 
We have demonstrated that our model, DAR-Net, can accurately and robustly analyze diver activities from underwater scenes by learning from both global and local context. 
A key contribution is our proposed multi-loss objective function, which integrates supervision from scene semantics to ensure the network focuses on the most relevant elements -- divers, robots, and objects of interest -- and ignores irrelevant background noise.
To facilitate research in this domain, we have curated and will make publicly available the first-ever Underwater Diver Activity (UDA) dataset. 
By providing over $2600$ semantically segmented images of diverse underwater human-robot collaborative scenarios, this dataset serves as a crucial foundation for future research in this challenging domain.
Through extensive experimental evaluations, DAR-Net has proven its effectiveness, consistently outperforming existing state-of-the-art models. 
While our framework shows promising results, particularly in its ability to focus on salient scene elements, future work will focus on expanding both the research scope and the UDA dataset, following the paths suggested in Sec.~\ref{sec:limitations}. 
We believe this framework and dataset will pave the way for more intelligent, proactive, and safe autonomous underwater vehicles in the future.

\bibliography{bib}
\bibliographystyle{IEEEtran}

\end{document}